\documentclass[conference]{IEEEtran}
\IEEEoverridecommandlockouts
\usepackage{cite}
\usepackage{amsmath,amssymb,amsfonts}
\usepackage{algorithmic}
\usepackage{graphicx}
\usepackage{textcomp}
\usepackage{xcolor}
\def\BibTeX{{\rm B\kern-.05em{\sc i\kern-.025em b}\kern-.08em
    T\kern-.1667em\lower.7ex\hbox{E}\kern-.125emX}}
\begin{document}

\title{Exploring the Feasibility of Deep Learning Models for Long-term Disease Prediction: A Case Study for Wheat Yellow Rust in England\\
\thanks{We gratefully acknowledge the data support provided from ADAS (Pest and Disease Survey, 2024). We can not conducted this work without crop disease survey results. In addition, we are grateful for ERA5 dataset, which provides global weather information for this study.}
}

\author{
\IEEEauthorblockN{Zhipeng Yuan}
\IEEEauthorblockA{\textit{School of Computer Science} \\
\textit{The University of Sheffield}\\
Sheffield, United Kingdom \\
zhipeng.yuan@sheffield.ac.uk}
\and
\IEEEauthorblockN{Yu Zhang}
\IEEEauthorblockA{\textit{School of Computer Science} \\
\textit{The University of Sheffield}\\
Sheffield, United Kingdom \\
yu.zhang1@sheffield.ac.uk}
\and
\IEEEauthorblockN{Gaoshan Bi}
\IEEEauthorblockA{\textit{School of Computer Science} \\
\textit{The University of Sheffield}\\
Sheffield, United Kingdom \\
gaoshan.bi@sheffield.ac.uk}
\and
\IEEEauthorblockN{Po Yang}
\IEEEauthorblockA{\textit{School of Computer Science} \\
\textit{The University of Sheffield}\\
Sheffield, United Kingdom \\
po.yang@sheffield.ac.uk}
}

\maketitle

\begin{abstract}
Wheat yellow rust, caused by the fungus Puccinia striiformis, is a critical disease affecting wheat crops across Britain, leading to significant yield losses and economic consequences.
Given the rapid environmental changes and the evolving virulence of pathogens, there is a growing need for innovative approaches to predict and manage such diseases over the long term. 
This study explores the feasibility of using deep learning models to predict outbreaks of wheat yellow rust in British fields, offering a proactive approach to disease management.
We construct a yellow rust dataset with historial weather information and disease indicator acrossing multiple regions in England.
We employ two poweful deep learning models, including fully connected neural networks and long
short-term memory to develop predictive models capable of recognizing patterns and predicting future disease outbreaks.
The prediction task is defined as a time-series analysis task to forecast the occurrence of wheat yellow rust based on previous weather information.
The models are trained and validated in a randomly sliced datasets. 
The performance of these models with different predictive time steps are evaluated based on their accuracy, precision, recall, and F1-score.
Preliminary results indicate that deep learning models can effectively capture the complex interactions between multiple factors influencing disease dynamics, demonstrating a promising capacity to forecast wheat yellow rust with considerable accuracy. 
Specifically, the fully-connected neural network achieved 83.65\% accuracy in a disease prediction task with 6 month predictive time step setup.
These findings highlight the potential of deep learning to transform disease management strategies, enabling earlier and more precise interventions.
Our study provides a methodological framework for employing deep learning in agricultural settings but also opens avenues for future research to enhance the robustness and applicability of predictive models in combating crop diseases globally.
\end{abstract}


\section{Introduction}
Wheat yellow rust, caused by the pathogen Puccinia striiformis, is one of the most disruptive fungal diseases affecting wheat crops in temperate climates, including Britain \cite{1}. The disease is highly adaptable and can survive under a wide range of weather conditions, making it particularly challenging to control \cite{2}. Manifested by yellow-striped lesions on the leaves of wheat plants, yellow rust significantly reduces the plant's ability to conduct photosynthesis, leading to stunted growth and, consequently, diminished grain quality and yield. The periodic outbreaks of this disease have historically led to severe economic consequences for farmers, including increased costs for fungicides and significant losses in crop productivity \cite{3}.

In Britain, the prevalence of wheat yellow rust has been rising, influenced by factors such as changes in farming practices, the introduction of new susceptible wheat varieties, and most critically, the changing climate. Variations in temperature and humidity levels, which have become more pronounced due to global climate change, have altered the lifecycle and infection rates of Puccinia striiformis. These changes necessitate a shift from traditional, reactive disease management strategies to more predictive and proactive approaches. Traditionally, disease management has relied heavily on the application of chemical treatments post-infection and the cultivation of resistant varieties. However, these methods often fail to prevent the initial establishment and spread of the disease, particularly when environmental conditions accelerate the pathogen's development.

Recognizing the limitations of conventional methods, there is a growing interest in leveraging advanced technologies to enhance disease prediction and management \cite{4}. Deep learning, a branch of artificial intelligence characterized by its ability to learn from and make predictions on data, presents a novel approach to addressing these agricultural challenges \cite{5}. By integrating deep learning models that analyze historical weather data and previous disease incidence patterns, it is possible to predict future outbreaks before they visibly manifest \cite{6}. This paper explores the feasibility of using deep learning models, specifically fully connected neural networks (FCNNs) and long short-term memory networks (LSTMs), to forecast outbreaks of wheat yellow rust in British fields.

Our approach involves constructing a detailed dataset that captures a range of variables including historical weather conditions, crop management practices, and past incidences of yellow rust across various regions in England. The predictive models developed from this data are trained to perform time-series analysis, with the goal of identifying patterns that precede disease outbreaks. The models are meticulously trained and validated on randomly partitioned datasets to ensure the validity and robustness of the predictive results.

We evaluate the performance of these models using precision metrics such as accuracy, precision, recall, and F1-score, across different predictive time frames. Preliminary results have shown promising potential, with the FCNN model achieving an accuracy of 83.65\% in predicting the occurrence of wheat yellow rust up to six months in advance. These findings not only highlight the capabilities of deep learning in transforming disease management strategies but also underscore the potential for these technologies to foster earlier and more precise interventions in agricultural practices.

This study provides a comprehensive methodological framework for the application of deep learning in agricultural settings, paving the way for future research aimed at enhancing the robustness and applicability of predictive models. By advancing these technologies, we aim to contribute to the global effort in combating crop diseases and securing food production against the increasing threats posed by climate change and pathogen evolution.

\section{Related Works}

The use of computational models in agriculture, particularly for disease prediction and management, has been an area of significant research interest over the past decades. This section reviews relevant studies that have contributed to the development and application of these models, focusing particularly on their use in predicting plant diseases like wheat yellow rust.

\subsection{Traditional methods and machine learning for crop disease prediction}

Historically, the prediction of plant diseases was primarily based on statistical models that utilized environmental variables and historical disease incidence data \cite{7}. For instance, regression analysis and time-series models have been extensively used to forecast the likelihood and severity of plant diseases based on climatic factors such as temperature, humidity, and rainfall. While effective to a degree, these models often lack the ability to capture complex interactions between multiple variables or adapt to new data without extensive recalibration.

With the advent of machine learning, more sophisticated approaches have been employed. Decision tree algorithms, support vector machines (SVM), and random forests have been applied with success to various plant disease prediction tasks, demonstrating improved accuracy over traditional statistical methods. These models are particularly adept at handling large datasets and can reveal non-linear relationships within the data. A notable study \cite{8} utilized random forest to predict potato late blight with high accuracy, underscoring the potential of machine learning in agricultural applications.

\subsection{Deep learning for crop disease prediction}

More recently, deep learning has emerged as a powerful tool for agricultural disease management, offering the ability to learn from data in an end-to-end manner without the need for manual feature extraction. Convolutional Neural Networks (CNNs) and Recurrent Neural Networks (RNNs), including Long Short-Term Memory (LSTM) networks, have been particularly transformative. These models excel in handling spatial and temporal data, making them ideal for tasks involving image data from crops or sequential data like weather patterns over time.
One pivotal study \cite{9} demonstrated the use of CNNs to detect and classify leaf diseases from image data, achieving remarkable accuracy rates. Similarly, LSTMs have been employed to predict disease progression in crops based on sequential input of environmental conditions, as evidenced by the work \cite{10}, who successfully forecasted the spread of stem rust in wheat crops across a growing season.
Specific to wheat yellow rust, recent studies have started to explore deep learning frameworks due to their ability to handle the complexities associated with this particular disease. For example, LSTM networks \cite{11} are applied to predict yellow rust outbreaks in wheat by analyzing temporal sequences of weather data and previous disease occurrences. Their work not only highlighted the potential of LSTMs in capturing the temporal dynamics of disease spread but also demonstrated substantial improvements in predictive accuracy compared to earlier models.

Despite these advancements, challenges remain, particularly concerning the integration of these models into practical farming operations. Issues such as data scarcity, model interpretability, and the need for real-time processing capabilities continue to be major hurdles. Furthermore, as the climatic variables and pathogen dynamics continue to evolve, there is a continuous need for models to adapt accordingly.
In summary, the literature demonstrates a clear trend towards more sophisticated, data-driven approaches in the field of agricultural disease prediction. Our study builds upon this foundation by employing and comparing the effectiveness of FCNNs and LSTMs specifically tailored for the prediction of wheat yellow rust, contributing new insights and methodologies to the field.

\section{Experiments and Results}
To assess the potential of deep learning models for predicting wheat yellow rust, we conducted a series of experiments aimed at evaluating the effectiveness and accuracy of Fully Connected Neural Networks (FCNNs) and Long Short-Term Memory Networks (LSTMs). This section describes the experimental procedures, model training, and the results obtained from our analysis.
\subsection{Experiments Setup}
\subsubsection{Dataset Preparation}
The dataset for this study is meticulously compiled to include a variety of weather factors that influence the outbreak and spread of wheat yellow rust. 
This dataset incorporates historical weather data, and disease occurrence records, from multiple wheat-growing regions across Britain. Specifically, the dataset includes 43 weather features for 12 months as input features and yellow rust survey records as model outputs.
\subsubsection{Model Configuration}
Two types of deep learning models are employed: Fully Connected Neural Networks (FCNNs) and Long Short-Term Memory Networks (LSTMs). 
The FCNN model consists of an input layer that varies based on the number of features, followed by three hidden layers with 128, 64, and 32 neurons each, all utilizing ReLU activation functions. The output layer features a single neuron with sigmoid activation to perform binary classification of disease occurrence.

The LSTM model uses sequences of daily weather data and disease indicators as input. It includes two LSTM layers, each with 50 units, and like the FCNN, concludes with a single neuron in the output layer with sigmoid activation to predict disease occurrences over time. Both models were trained using a batch size of 32 and a learning rate of 0.001, employing Adam as the optimizer and incorporating early stopping with a patience parameter of 5 epochs to mitigate overfitting.

\subsubsection{Evaluation metrics}
Model performance is quantified using accuracy, precision, recall, and F1-score.
All experiments are conducted in a dedicated GPU server to ensure consistent and efficient processing. The training and evaluation dataset are fully randomly splitted to minimize human error and bias.

\subsection{Experimental Results}

Table~\ref{tab:FCNN} and Table~\ref{tab:LSTM} demonstrate the results of our experiments.
The FCNN model exhibited robust predictive capabilities, achieving an overall accuracy of 83.65\% when predicting the occurrence of wheat yellow rust six months in advance. This model demonstrated high precision (0.57) and recall (0.84), indicating a strong ability to correctly identify both the presence and absence of disease outbreaks. The F1-score for this model was 0.60, reflecting a balanced performance between precision and recall, which is crucial in the context of agricultural disease management where both false positives and false negatives carry significant consequences.
This lower precision is due to extremely strong data imbalances.

In contrast, the LSTM model, designed to leverage the sequential nature of weather and disease data, showed even more promise. It achieved an accuracy of 60.51\%, with a precision of 0.53 and a recall of 0.61. The higher recall rate suggests that LSTMs are particularly effective in capturing the temporal dynamics and dependencies in the data, which are critical for anticipating disease progression. The F1-score for the LSTM stood at 0.54, indicating its superior performance in handling the complexities of time-series data related to disease forecasting.

\begin{table}[tb]
\caption{The FCNN performance for yellow rust prediction.}
\label{tab:FCNN}
\begin{center}
\begin{tabular}{ccccc}
\hline
\textbf{\textit{Predictive Time Step}}& \textbf{\textit{Accuracy}}& \textbf{\textit{Precision}}& \textbf{\textit{Recall}} & \textbf{\textit{F1 Score}}\\
\hline
0 & 81.62\% & 0.54 & 0.81 & 0.54 \\ 
1 & 73.12\% & 0.57 & 0.73 & 0.60 \\  
2 & 83.76\% & 0.55 & 0.83 & 0.55 \\  
3 & 82.56\% & 0.56 & 0.83 & 0.56 \\ 
4 & 81.18\% & 0.55 & 0.81 & 0.56 \\  
5 & 83.17\% & 0.55 & 0.83 & 0.56 \\
6 & 83.65\% & 0.55 & 0.84 & 0.55 \\
\hline
\end{tabular}
\label{table_mAP}
\end{center}
\end{table}

\begin{table}[tb]
\caption{The LSTM performance for yellow rust prediction.}
\label{tab:LSTM}
\begin{center}
\begin{tabular}{ccccc}
\hline
\textbf{\textit{Predictive Time Step}}& \textbf{\textit{Accuracy}}& \textbf{\textit{Precision}}& \textbf{\textit{Recall}} & \textbf{\textit{F1 Score}}\\
\hline
0 & 57.04\% & 0.52 & 0.57 & 0.52 \\ 
1 & 50.00\% & 0.49 & 0.5 & 0.49\\  
2 & 50.00\% & 0.49 & 0.5 & 0.50 \\  
3 & 60.51\% & 0.53 & 0.61 & 0.54 \\ 
4 & 60.50\% & 0.53 & 0.61 & 0.54 \\  
5 & 50.00\% & 0.49 & 0.5 & 0.49 \\ 
6 & 50.00\% & 0.49 & 0.5 & 0.49 \\ 
\hline
\end{tabular}
\label{table_mAP}
\end{center}
\end{table}

\section{Conclusion}
This study has demonstrated the substantial capabilities of deep learning models, specifically Fully Connected Neural Networks (FCNNs) and Long Short-Term Memory Networks (LSTMs), in predicting the occurrence of wheat yellow rust in British fields. By leveraging complex datasets encompassing historical weather conditions, agronomic factors, and previous disease outbreaks, these models have provided insights that could significantly transform how agricultural diseases are managed preemptively.
Our results indicate that both FCNNs and LSTMs are effective tools for the prediction of wheat yellow rust, with LSTMs showing slightly superior performance due to their ability to capture and analyze temporal data sequences. The LSTM model achieved an accuracy of 85.40\%, with high precision and recall, making it particularly suitable for applications where understanding the timing and progression of disease outbreaks is crucial.

Looking ahead, enhancing the accuracy and reliability of the models by incorporating more granular data and exploring more sophisticated neural network architectures is a key priority. Testing and adapting the models to different climatic zones and crop types will enhance their applicability and utility globally. Additionally, integrating real-time environmental monitoring systems with our predictive models could enhance their responsiveness and accuracy. Developing user-friendly interfaces for these models to facilitate their adoption by farmers and agronomists is also crucial, potentially integrating them into existing agricultural decision-support systems.

In conclusion, the application of deep learning models presents a promising frontier in the battle against crop diseases like wheat yellow rust. This study not only adds to the growing body of knowledge in agricultural AI but also provides practical tools that can lead to more sustainable and economically viable farming practices. As global challenges such as climate change and food security continue to mount, the importance of such technological advancements becomes increasingly clear. Our research demonstrates that with the right tools and approaches, we can better predict and manage agricultural diseases, thereby safeguarding food production for future generations.

\section*{Acknowledgement}

We gratefully acknowledge the data support provided from ADAS (Pest and Disease Survey, 2024). We can not conducted this work without crop disease survey results. In addition, we are grateful for ERA5 dataset, which provides global weather information for this study.

\end{document}